%% file: main.tex
\documentclass[]{fairmeta}

 \setcitestyle{numbers,square}
 \usepackage[table]{xcolor}
 \usepackage{tabularx}
 \usepackage{amsmath}

\usepackage{booktabs}
\usepackage{siunitx}

\definecolor{ourrow}{HTML}{EEF3FB}
\definecolor{rulegray}{HTML}{555555}
 
\newcommand{\best}[1]{\textbf{#1}}
\newcommand{\second}[1]{\underline{#1}}
\newcommand{\gr}[1]{\textit{\textcolor{rulegray}{#1}}}
\newcommand{\sub}[1]{\hspace{1em}#1}   

\usepackage[utf8]{inputenc} 
\usepackage[T1]{fontenc}    
\usepackage{hyperref}       
\usepackage{url}            
\usepackage{booktabs}       
\usepackage{amsfonts}       
\usepackage{nicefrac}       
\usepackage{microtype}      
\usepackage{xcolor}         
\usepackage{graphicx}
\usepackage{amssymb}

\title{EV-JEPA: Elastic V-JEPA for Any Device and Cross-Device Inference}

\author[1,*]{Anxhelo Diko}
\author[1,*]{Jieyu Lin}
\author[1]{Stefano Ambrogio}
\author[1]{Ziyun Li}
\author[2]{Mahmoud Assran}
\author[2]{Théo Moutakanni}
\author[1]{Barbara De Salvo}
\author[1]{Chiao Liu}

\affiliation[1]{Meta Reality Labs Research}
\affiliation[2]{FAIR at Meta}

\contribution[*]{Equal contribution.}
\date{\today}

\abstract{Foundation models like V-JEPA are increasingly deployed across a heterogeneous edge–cloud continuum spanning smart glasses, phones, robots, and servers. Yet, they are typically trained as rigid, fixed-capacity networks bound to one device, supporting neither reconfigurable compute budgets nor collaborative execution. Existing approaches address only one axis at a time: slimmable and NAS-based methods provide width elasticity but assume single-device inference, whereas split-computing methods enable cross-device collaboration only for fixed partitions and static capacities. We introduce Elastic V-JEPA (EV-JEPA), a single model trained once and deployable everywhere along the continuum. Its width is reconfigurable, and it is natively partitionable across heterogeneous devices. 
 At its core is Transition-Aware Triplet Training (T3). At each step we pick a narrow and a wide width, and train three paths: each width alone, plus a hybrid path that runs the early layers narrow and the rest wide. The hybrid switch point is where one device hands off to another, so the same weights learn both single-device and cross-device execution.
 Across various benchmarks, EV-JEPA falls within 1 accuracy point of independently trained baselines while improving over previous elastic training approaches by up to 11 accuracy points. Narrow-to-wide cross-device hybrid execution recovers to within 1.1 points of the highest-capacity standalone model and supports many-to-one, any-to-any, and variable-split configurations. 
 Finally, we show that the elastic principle extends beyond the backbone to an elastic token compression mechanism that exposes controllable trade-offs among on-device computation, latency, and accuracy.
 }

\begin{document}

\maketitle

\input{Chapters/Introduction}

\input{Chapters/Related_works}
\input{Chapters/Method}

\input{Chapters/Experiments}
\input{Chapters/Conclusions}

\bibliographystyle{abbrvnat}
\bibliography{main}

\end{document}

%% file: Chapters/Introduction.tex
\section{Introduction}

Vision foundation models (VFMs) like V-JEPA~\cite{bardes2024vjepa} are a cornerstone of modern computer vision and AI, powering applications that range from classification~\cite{assran2025vjepa,bardes2024vjepa} to action anticipation and planning~\cite{assran2025vjepa,diko2024semantically}. Traditionally, these models are designed to run on powerful accelerators. However, the rise of wearable smart glasses~\cite{engel2023project} and robotics~\cite{shah2023gnm} is reshaping where such models must operate. Rather than relying on a centralized cloud, the deployment infrastructure is evolving into a distributed edge-cloud continuum \cite{zhou2019edge,dustdar2022distributed}. A heterogeneous, multi-tier architecture where edge devices (e.g. smart glasses and robots) with different compute budgets and remote servers cooperate to serve inference. 
Conventionally, such cooperation has involved edge devices streaming raw video to the cloud for processing, which is at odds with the need for smart glasses and robots to assist and act in the physical world in real time. Streaming is power-hungry and slow due to low-bandwidth wireless links the edge relies on~\cite{dong2022splitnets}, making transmission a primary bottleneck.
These constraints motivate research on efficient VFMs~\cite{mehta2021mobilevit,howard2017mobilenets,tan2019efficientnet,maes2603leworldmodel}, promoting on-device compute. 
Yet, these approaches share two key limitations: i) they train a separate model per compute budget, and ii) they treat each device in isolation, so an edge device is capped at its own budget and the collaborative compute still relies on video streaming.

To address i), one line of work makes a single network serve multiple compute budgets, whether by extracting sub-architectures from a supernetwork~\cite{cai2020once} or through variable widths \cite{devvrit2024matformer,zhang2024slicing,hou2020dynabert}. These methods differ in how far elasticity reaches, from a single sublayer's width~\cite{devvrit2024matformer} to the hidden dimension itself~\cite{zhang2024slicing}, but not in what they produce: a set of self-contained models (one per compute budget) whose representations are not cross-width compatible. They are intended to serve different individual devices, but not to split a single inference across them.
A separate line of work addresses ii) by splitting the network’s workload across multiple heterogeneous compute nodes~\cite{dong2022splitnets}. While effective for exploiting distributed hardware, they assume a fixed partition and a static per-node capacity, forfeiting standalone execution and treating distributed inference as a requirement rather than an option. As a result, existing approaches either provide width elasticity or collaboration, but not both. The two failures interlock as elastic methods cannot carry a representation across widths, and split methods can serve only rigid architectures. 
This raises a fundamental question: can we train neural networks that are both elastic and collaborative, and be able to run fully on a single device and seamlessly split their computation across multiple ones when needed? 

\begin{figure}[!t]
    \centering
    \includegraphics[width=\linewidth]{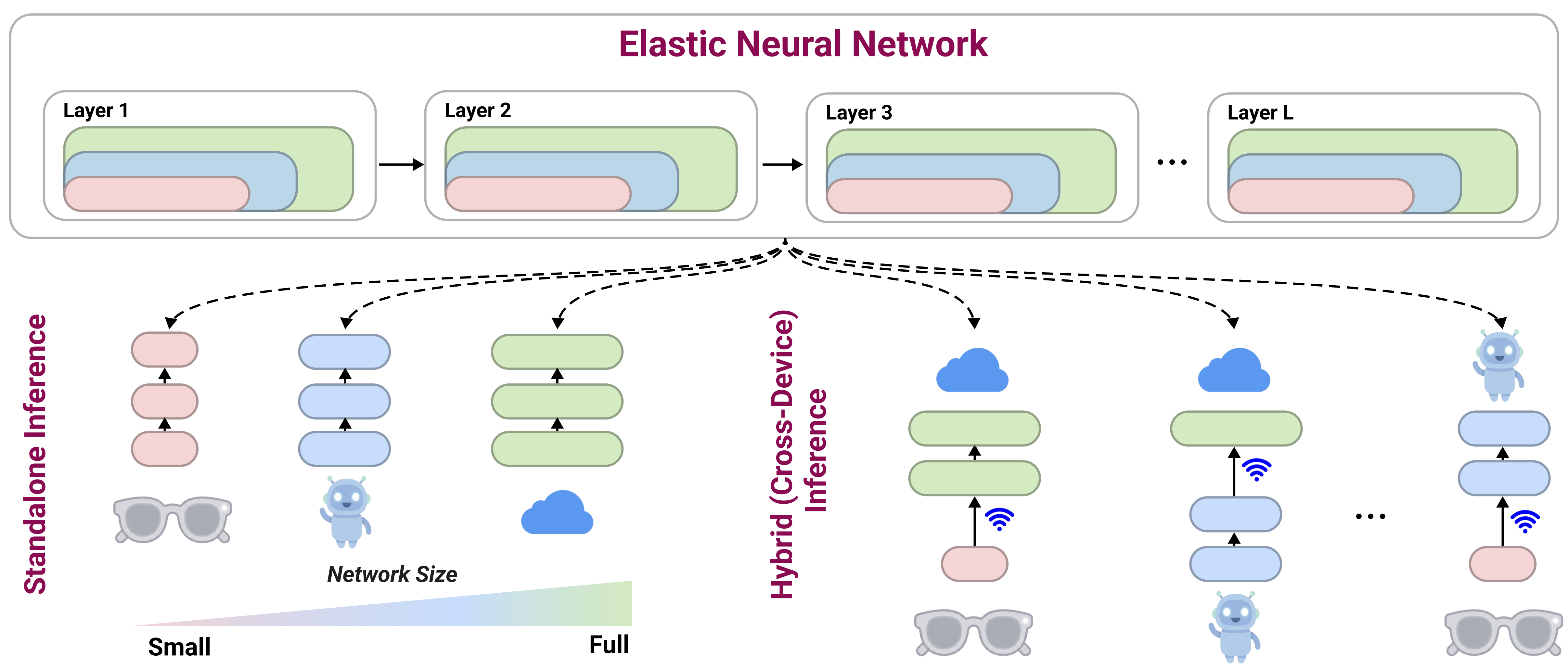}
    \caption{%
        \textbf{One elastic model for the device continuum.}
        We propose a single \emph{elastic network} that is trained once and deployed across the entire continuum: (1)~its width is reconfigurable at inference time, allowing the same parameters to serve any compute budget \textbf{\textit{(Standalone Inference)}}; and (2)~its computation can be \emph{split} between devices, with early layers executed on-device and remaining layers on a stronger device, communicating only compact intermediate representations \textbf{\textit{(Hybrid Inference)}}.%
    }
    \label{fig:teaser}
\end{figure}

To answer this question, we introduce a novel approach that achieves both width elasticity and collaborative execution instantiated on the self-supervised predictive V-JEPA~\cite{bardes2024vjepa} model named as \textbf{Elastic V-JEPA (EV-JEPA)} (Fig.~\ref{fig:teaser}). EV-JEPA builds on a vision transformer backbone and supports full-width elasticity across all layer components (Fig.~\ref{fig:teaser} \textbf{Standalone Inference}). In addition, beyond standalone width elasticity, we depart from previous works by introducing a  \textit{hybrid} (mixed-width) execution mode in which early layers run at a smaller width and the remainder at a larger one, transitioning at a variable split layer. This allows a small device to initiate inference and a more capable one to complete it, thereby improving inference accuracy by streaming tokens rather than videos. However, it imposes a requirement that no elastic method has had to satisfy: representations that are cross-width compatible. 

At the core of our approach is a novel training paradigm we term Transition-Aware Triplet Training (T3). Unlike prior elastic approaches that optimize a randomly selected width~\cite{devvrit2024matformer} or rigid \textit{\{smallest, random, largest\}} triplet \cite{yu2019universally,yu2019slimmable}, often resulting in performance degradation at intermediate capacities, T3 continuously samples a \textit{\{smaller, hybrid, larger\}} triplet. 
The same width pair is therefore optimized simultaneously as two standalone configurations, and as the endpoints of the hybrid path that connects them, so cross-width compatibility is trained directly as a native capability. Additionally, T3 builds on the same training frameworks as the original V-JEPA with one additional twist: the self-evolving EMA teacher is replaced with a fixed pretrained V-JEPA 2~\cite{assran2025vjepa} encoder to provide a stable target space across all widths. This means EV-JEPA is trained with a dense predictive objective where the student must estimate teacher representations per token, an objective significantly stricter than agreeing on a global representation (i.e., classification)~\cite{yu2019universally,devvrit2024matformer,zhang2024slicing}, making elasticity even harder. To our knowledge, this is the first elastic framework proposed on a densely supervised model.
Finally, to close the gap between a network that can be split in principle and one that runs efficiently over a device continuum connected through low-bandwidth links, we also experiment coupling hybrid execution with an elastic, controllable token-compression layer to ease the cross-device communication.



We extensively evaluate our elastic-collaborative framework on video understanding benchmarks, including Something-Something V2~\cite{goyal2017something} and Kinetics-400~\cite{kay2017kinetics}, as well as the predictive task of action anticipation on Epic-Kitchens 100~\cite{damen2022rescaling}.
Our empirical results demonstrate that EV-JEPA trained with T3 achieves superior performance compared to other training regimes and also resolves the intermediate subnetwork degradation~\cite{zhang2024slicing,yu2019universally}.
In summary, our contributions are threefold.
First, we introduce \textbf{EV-JEPA}, an elastic VFM whose width is reconfigurable across all layer components and that natively supports hybrid mixed-width execution.
Second, we propose \textbf{T3} which optimizes a width pair simultaneously as two standalone configurations and as the endpoints of the hybrid path connecting them, making cross-width compatibility a trained property rather than an emergent one.
Third, we show across multiple benchmarks that EV-JEPA stays within 1 accuracy point of independently trained standalone baselines while improving over prior elastic training by up to 11 accuracy points, and that narrow-to-wide hybrid execution recovers to within 1.1 accuracy points of the highest-capacity under different configurations. We further show that the elastic principle extends to a token compressor, reducing intermediate communication, transfers to vision-language understanding under VL-JEPA\cite{chen2025vl} setting, and applies to convolutional backbones to support deployment and partition across compact, low-memory devices.

%% file: Chapters/Related_works.tex
\section{Related Works}


\noindent\textbf{Elastic Networks.} A first family of approaches supporting heterogeneous deployment stems from one-shot NAS. OFA~\citep{cai2020once} and BigNAS~\citep{yu2020bignas} train large convolutional supernets that treat width, depth, kernel size, and resolution as elastic dimensions, from which specialized sub-architectures are extracted after training; subsequent works refine training and search procedures~\citep{guo2020singlepath,wang2021attentivenas}.
While powerful, these methods frame elasticity as a search problem, and as the gap between the smallest and largest deployable configurations grows, the resulting search space expands combinatorially, making sub-architectures searching prohibitive.
Slimmable approaches emerged as a complementary alternative that trades search for a fixed family of nested widths sharing weights.
US-Net~\citep{yu2019slimmable,yu2019universally} introduced this idea in the convolutional setting by jointly training a set of fixed widths with switchable batch normalization, the sandwich training rule, and in-place distillation, becoming the standard recipe for slimmable training. These ideas were later transferred to transformers, first by DynaBERT~\citep{hou2020dynabert}, which adapts both width and depth of BERT through a two-stage distillation procedure, and then to vision transformers (ViT)~\citep{haberer2024hydravit,zhang2024slicing}, led by MatFormer~\citep{devvrit2024matformer}, which applies the matryoshka principle~\cite{kusupati2022matryoshka} to FFN blocks, training a fixed set of nested dimensions by sampling one granularity per step under a tunable probability distribution.
A third line of work bridges these two families by combining weight-sharing supernets with NAS-style search over transformer architectures~\citep{chen2021autoformer, chen2021searching,gong2022nasvit}.
Despite their progress, all of these methods share two limitations relevant to our setting: 1) degenerated performance in intermediate widths~\citep{cao2023three}, and 2) networks are designed to run on isolated hardware. Our work directly addresses both, producing a single elastic model whose width is reconfigurable across all layer components and that natively supports hybrid mixed-width execution that splits compute from a smaller to a bigger device.  \\
\noindent\textbf{Split Computing.} A complementary line of work addresses distributed inference by partitioning a network across devices.
Neurosurgeon~\citep{kang2017neurosurgeon} pioneered layer-granularity splitting between mobile and cloud, selecting the partition point at runtime based on network conditions and device load.
SPINN~\citep{laskaridis2020spinn} extended this idea by combining early-exit mechanisms with device-cloud splitting, co-optimizing both policies at runtime for latency-constrained settings.
SplitNets~\citep{dong2022splitnets} specifically targeted head-mounted systems, designing neural architectures for efficient distributed computation across wearable sensors and edge servers.
While these methods demonstrate the value of collaborative execution, they operate on fixed architectures running only in distributed mode. Our work bridges these two lines by producing a single model that is both elastic and natively splits compute, trained end-to-end to support standalone and collaborative inference across heterogeneous devices.

%% file: Chapters/Method.tex
\section{Method}
\label{s:method}

\subsection{Problem Formulation}
\label{ss:problem_formulation}

We consider the setting of \emph{elastic \& distributed inference}, where a model operates across multiple compute capacities and optionally distributes computation across devices. Let $f_\theta$ denote a model parameterized by $\theta$, and $L$ be the number of layers in the model.  Instead of a fixed architecture, we assume the model supports a set of \emph{width configurations} $w \in \mathcal{W}$ where $\mathcal{W}$ is a predefined set of supported widths corresponding to different compute budgets (e.g., $\mathcal{W}=\{192,384,\dots,1408\}$). For a given width $w$, we denote the corresponding subnetwork as:
\begin{equation}
f_\theta^{(w)}(x),
\end{equation}
which represents the model executed entirely at width $w$. Elastic inference requires that a single set of parameters $\theta$ performs well across all configurations $w \in \mathcal{W}$. In distributed deployments, inference may also be \emph{split across devices}. A resource-constrained device may execute an initial portion of the model using width $w_s$, after which the intermediate representation is transferred to a stronger device that continues computation using a larger width $w_l$, with $w_s < w_l$. We therefore define a \emph{hybrid configuration}
\begin{equation}
f_\theta^{(w_s, w_l, k)}(x),
\end{equation}
where the first $k$ layers operate with width $w_s$ and the remaining $L-k$ layers operate with width $w_l$. 
The goal of our training framework is therefore twofold. First, the model should achieve strong performance across all standalone configurations $f_\theta^{(w)}$, enabling deployment on devices with different compute budgets. Second, the model should support \emph{distributed execution}, where computation can be distributed across devices. 
%

\begin{figure}[t]
    \centering
    \includegraphics[width=\linewidth]{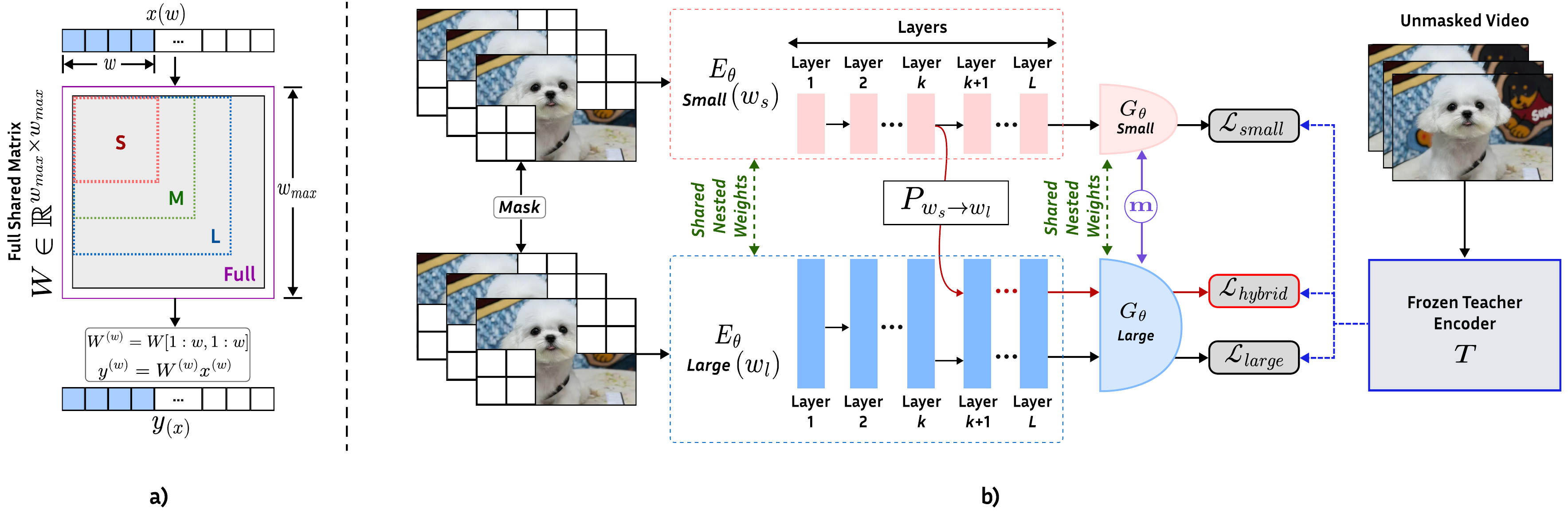}
    \caption{\textbf{Elastic architecture and Transition-Aware Triplet Training (T3).}
    \textbf{(a)} Weight slicing. All width configurations share a single set of parameters:
    for a layer weight $W \in \mathbb{R}^{w_{\max}\times w_{\max}}$, a width-$w$ subnetwork uses the top-left slice $W^{(w)} = W[1{:}w, 1{:}w]$, so all configurations are nested submatrices of the same $W$. \textbf{(b)} T3 training. At each iteration we sample a width pair $(w_s, w_l)$ and a split layer $k$, and jointly optimize three paths over the \emph{same} masked input: a \textbf{small} standalone encoder $E_\theta (w_s)$, a \textbf{large} standalone encoder $E_\theta (w_l)$, and a hybrid path that runs the first $k$ layers at $w_s$ and transitions to $w_l$ at layer $k$ via the elastic operator $P_{w_s\rightarrow w_l}$. Each encoder feeds a predictor $G_\theta$ together with a learned mask token $\mathbf{m}$; the three paths produc $\mathcal{L}_{\text{small}}$, $\mathcal{L}_{\text{hybrid}}$, and
    $\mathcal{L}_{\text{large}}$, all regressed against targets from a frozen V-JEPA teacher
    $T$ on the unmasked video. The predictor width is set by a capped rule
    $w_G = \min(w_G^{\max}, \tfrac{1}{2} w_E)$ (Eq.~\ref{eq:predictor-width}) and is therefore \emph{not} tied to the encoder width; the labels ``Small''/``Large'' denote which path a predictor serves, not its capacity.}
    \label{fig:method_v2}
\end{figure}

\subsection{Elastic Transformer Architecture }
\label{ss:elastic_arch}
We design a transformer architecture whose hidden dimension can vary across a predefined set of widths $\mathcal{W}$. Each width configuration corresponds to a subnetwork obtained by selecting the first $w$ channels of the model parameters.
Concretely, following slimmable networks~\cite{yu2019slimmable} and Matformer~\cite{devvrit2024matformer}, the parameters of the largest model with width $w_{\max}$ are shared across all configurations through weight slicing. Consider a transformer layer with attention projections $W_Q, W_K, W_V \in \mathbb{R}^{w_{\max} \times w_{\max}}$. As shown in Fig.~\ref{fig:method_v2} (a), a smaller configuration with $w < w_{\max}$ uses the sliced parameters:
\begin{equation}\label{eq:w_slize}
W_Q^{(w)} = W_Q[1{:}w,1{:}w], \quad
W_K^{(w)} = W_K[1{:}w,1{:}w], \quad
W_V^{(w)} = W_V[1{:}w,1{:}w].
\end{equation}

The same principle applies to the feed-forward network. Given the full FFN parameters $W_1 \in \mathbb{R}^{r w_{\max} \times w_{\max}}$ and $W_2 \in \mathbb{R}^{w_{\max} \times r w_{\max}}$, with $r$ being the upscaling factor, a width-$w$ uses the slices:
\begin{equation}\label{eq:w_slice_2}
W_1^{(w)} = W_1[1{:}rw,1{:}w], \quad
W_2^{(w)} = W_2[1{:}w,1{:}rw].
\end{equation}
This nested parameterization allows all subnetworks to reuse a single set of weights while maintaining architectural compatibility across widths. The supported $\mathcal{W}$ form a discrete set due to the constraint that the hidden dimension must remain divisible by the attention head size.

For hybrid configuration where smaller earlier layers are connect to larger later layers, the architecture supports width transitions between layers. When computation transitions from width $w_s$ to a larger width $w_l$ at layer $k$, the intermediate representation $h^{(w_s)}_{k} \in \mathbb{R}^{w_s}$ is expanded to $\mathbb{R}^{w_l}$ through a learned linear projection $P$. The operator is itself elastic: we maintain a single matrix $P \in \mathbb{R}^{w_{\max} \times w_{\max}}$, and for any transition $(w_s, w_l)$ we use the slice:
\begin{equation}\label{eq:projection_slicing}
    P_{w_s \to w_l} = P[1{:}w_l,\ 1{:}w_s] \in \mathbb{R}^{w_l \times w_s}
\end{equation}
%
so that $h^{(w_l)}_k = P_{w_s \to w_l}\, (h^{(w_s)})$. This nested parameterization mirrors the slicing scheme of the transformer blocks where a single set of projection parameters is shared across all $(w_s, w_l)$ pairs. The same $P$ is further reused across all possible split layers $k \in \{2, \dots, L\}$, making the operator invariant to where the transition occurs. This design yields a single compact transition module shared across the entire elastic configuration space and split layers $k$.

\subsection{Transition-Aware Triplet Training}
\label{ss:pts}


Our architecture introduces a capability absent from prior elastic methods: \emph{hybrid execution}, where computation transitions from width $w_s$ to a larger width $w_l$ at a split layer $k$\footnote{Note that this capability is different from mix and match from Matformer\cite{devvrit2024matformer}. Mix and match works because the input and output dims remain unchanged in the original Matformer.}. This capability is orthogonal to standalone elasticity. Our objective is therefore to train a single model that performs well both at any standalone width $w \in \mathcal{W}$ and across hybrid configurations $(w_s, w_l, k)$. We formalize this by extending the configuration space. Traditional elastic training optimizes standalone subnetworks sampled from the width set $\mathcal{W}$:
\begin{equation}\label{eq:training_standalone}
    \min_\theta \; \mathbb{E}_{w \sim p(\mathcal{W})} \; \mathcal{L}\!\left(f_\theta^{(w)}\right).
\end{equation}
In contrast, our formulation expands the configuration space to include hybrid executions,
\[
\mathcal{C} = \mathcal{W} \cup \{(w_s, w_l, k) : w_s, w_l \in \mathcal{W},\ w_s < w_l,\ k \in \{2,\dots,L-1\}\},
\]
and optimizes configurations drawn from this joint space:
\begin{equation}\label{eq:training_triplet}
\min_\theta \; \mathbb{E}_{c \sim p(\mathcal{C})} \; \mathcal{L}\!\left(f_\theta^{(c)}\right).
\end{equation}
In this formulation, hybrid execution $f^{(w_s,w_l,k)}_\theta$ is treated as a first-class configuration rather than as an emergent property of training.
This naturally allows existing sampling strategies to be extended to the hybrid setting. We consider two such extensions as baselines:

\begin{itemize}
\item \textbf{Distributional sampling (\textit{Matformer-style})}~\cite{devvrit2024matformer}: a configuration $c \sim p(\mathcal{C})$ is sampled independently at each iteration. While this provides wide coverage of the configuration space, configurations are optimized independently.

\item \textbf{Sandwich rule (\textit{US-Net-style)}}~\cite{yu2019universally}: the smallest and largest widths are optimized together with one random configuration $c_{\text{rand}} \sim p(\mathcal{C})$ per iteration. This provides strong supervision at the boundaries of the elastic space, but intermediate widths $w$ (i.e, $w_{\min} < w < w_{\max}$) now compete with hybrid configurations for the same sampling slot.
\end{itemize}

As a more suitable strategy for the joint standalone and hybrid objective, we propose \textbf{Transition-Aware Triplet Training (T3)}. At each iteration, we sample:
\[
w_s, w_l \sim \mathcal{W},\quad w_s < w_l, 
\qquad
k \sim \{2,\dots,L-1\},
\]
and optimize the triplet:
\begin{equation}\label{eq:T3_triplet}
\mathcal{T}_{T3} =
\{ f^{(w_s)}_\theta,\ f^{(w_s, w_l, k)}_\theta,\ f^{(w_l)}_\theta \},
\end{equation}
which mirrors the three deployment modes: standalone small, hybrid execution, and standalone large.
T3 differs from the extended baselines in one key respect: the same width pair $(w_s, w_l)$ appears simultaneously as standalone configurations and as the endpoints of a hybrid path within the same iteration. This couples the three configurations through \textbf{shared gradients}, directly optimizing both standalone performance and cross-width compatibility, rather than relying on these to emerge from independent sampling.

\subsection{Training}
\label{ss:vjepa_distillation}


We train the elastic model under a V-JEPA-style masked latent prediction objective~\cite{assran2025vjepa, bardes2024vjepa}. 
V-JEPA follows an encoder--predictor regime in which a student model observes a masked video and predicts latent representations for the masked regions from the visible context using the predictor and a learned mask token $\mathbf{m}$\footnote{Check the appendix for more V-JEPA details}. 
In our setting, we retain this predictive training structure but replace the original teacher mechanism with distillation from a frozen pretrained V-JEPA model. 
Concretely, the student receives a masked input video, predicts latent embeddings for the masked tokens, and is optimized to match target representations produced by the frozen teacher on the corresponding full video.

Given an input video $v$, it is first split into a sequence of spatio-temporal
tokens $X = \{x_1,\dots,x_N\}$.
A subset of tokens is masked according to the multi-block masking strategy
introduced in V-JEPA~\cite{bardes2024vjepa}.
Specifically, we sample spatially contiguous blocks and extend them across the
full temporal extent of the clip, producing tubelet masks whose union defines the
masked index set $\mathcal{M} \subset \{1,\dots,N\}$ and its complement
$\mathcal{V} \subset \{1,\dots,N\} \setminus \mathcal{M}$ indexes the visible (unmasked) tokens.
For an index set $\mathcal{V}$ we write
$X_{\mathcal{V}} = \{x_i\}_{i \in \mathcal{V}}$.
$X_{\mathcal{V}}$ are passed to the student encoder
$E_\theta$, while the predictor $G_\theta$ predicts latent representations for
the masked positions $i \in \mathcal{M}$.
This training setup is shared across all configurations $c \in \mathcal{C}$. 
In particular, the same $(X,M)$ pair is used for all configurations in a T3 triplet so that standalone and hybrid paths are exposed to the same predictive task.

%

\paragraph{Student.} The student is our parametrized learnable model $f_\theta$, a composition of our learnable modules including the encoder $E_\theta$ and the predictor $G_\theta$, both built on the elastic transformer architecture according to the weight-sharing policy defined in Section~\ref{ss:elastic_arch}. A sampled configuration $c \in \mathcal{C}$ determines how the encoder is executed. A standalone configuration is specified by a single width $w \in \mathcal{W}$, with all
encoder layers run at width $w$. A hybrid configuration is specified by a tuple
$(w_s, w_l, k)$, where the first $k$ encoder layers run at width $w_s$ and the remaining
$L-k$ layers run at width $w_l$. In both cases, we write $w_E(c)$ for the encoder
output width, where $w_E(c) = w$ for standalone configurations and $w_E(c) = w_l$ for
hybrid configurations. Then, the predictor's elastic width is derived from the $w_E(c)$ by the following rule:
\begin{equation}\label{eq:predictor-width}
    w_G(c) = \min\!\left(w_G^{\max},\ \tfrac{1}{2}\, w_E(c)\right),
    \qquad w_G^{\max} = 384,
\end{equation}
so the predictor tracks half the encoder width up to a cap $w_G^{\max}$. The cap $w_G^{\max} = 384$ matches the predictor width of the original V-JEPA~\cite{assran2025vjepa,bardes2024vjepa} while the half-width rule scales the predictor \emph{below} this reference for smaller encoders, deliberately limiting predictor capacity so that the encoder is forced to carry the representational load. Once the widths are defined, for a sampled configuration $c \in \mathcal{C}$, the student prediction for a masked token $i \in M$ is:
%
\begin{equation}\label{eq:prediction}
    z_{S,i}^{(c)} = \left[f_\theta^{(c)}(X_\mathcal{V}) \right]_i = \left[ G_\theta^{(w_G(c))}\!\left(
        W_{\text{proj}}^{(w_E(c)\rightarrow w_G(c))}\, \left(E_\theta^{(w_E(c))}(X_\mathcal{V})\right),\
        \mathbf{m}
    \right) \right]_i,
\end{equation}
where $W_{\text{proj}}^{(w_E(c)\rightarrow w_G(c))}$ is an elastic linear projection that maps the encoder output to $w_G(c)$, and $\mathbf{m} \in \mathbb{R}^{w_G(c)}$ is the learned mask token inserted at the masked positions $i \in M$. Both $W_{\text{proj}}$ and $\mathbf{m}$ are sliced under the same weight-sharing policy as Section~\ref{ss:elastic_arch}, so the predictor input is fully defined at width $w_G(c)$ for every configuration $c$. \\

\noindent\textbf{Teacher.} Target representations are produced by a frozen, pretrained V-JEPA~2 ViT-G teacher~\cite{assran2025vjepa}, denoted $T$. Unlike the original V-JEPA formulation, where the teacher is updated via an exponential moving average (EMA) of the student, we keep $T$ fixed throughout training. An EMA teacher is ill-suited to the elastic setting: since the student is sampled at a different configuration $c \in \mathcal{C}$ at each iteration, an EMA target would average parameters across configurations of varying capacity, yielding an unstable and width-dependent learning signal. A frozen full-capacity teacher instead provides a stable target shared across all configurations, anchoring every subnetwork to the same representation space. Exploring EMA-style updates in the elastic regime is beyond the scope of this work.
Given the full (unmasked) input, the teacher produces target embeddings
\begin{equation}\label{eq:teacher_pred}
    z_{T,i} = T(X)_i, \quad i \in M,
\end{equation}
for the masked positions. Because the teacher operates at a fixed hidden dimension that differs in general from the student's configuration-dependent output dimension, we align student and teacher representations through a lightweight learned elastic head $h_\phi$ applied to the student predictions. For notational simplicity, we fold $h_\phi$ into $z_{S,i}^{(c)}$ in what follows. \\
\noindent\textbf{Loss.} Following V-JEPA, we use an $\ell_1$ regression loss on the masked tokens:
\begin{equation}\label{eq:l1}
\mathcal{L}(c) \;=\; \frac{1}{|M|} \sum_{i \in M} \left\| z_{S,i}^{(c)} - z_{T,i} \right\|_1 .
\end{equation}
This objective is applied to each of the configurations in the triplet $\mathcal{T}_{T3} = \{f^{(w_s)}_\theta, f^{(w_s, w_l, k)}_\theta, f^{(w_l)}_\theta\}$, giving
\begin{equation}\label{eq:loss_items}
\mathcal{L}_{\text{small}} = \mathcal{L}(w_s), \qquad
\mathcal{L}_{\text{hybrid}} = \mathcal{L}\!\left((w_s, w_l, k)\right), \qquad
\mathcal{L}_{\text{large}} = \mathcal{L}(w_l),
\end{equation}
and the total training loss is optimized jointly per iteration:
\begin{equation}
    \mathcal{L}_{\text{total}} = \mathcal{L}_{\text{small}} + \mathcal{L}_{\text{hybrid}} + \mathcal{L}_{\text{large}} .
\end{equation}
Since the teacher $T$ is frozen, all three configurations are regressed toward a single consistent target, which stabilizes elastic training and prevents divergence between subnetworks observed under distributional sampling (Section~\ref{ss:pts}).

%% file: Chapters/Experiments.tex
\begin{table}[!t]
\centering
\caption{\textbf{Elastic Model Setup.}
In the default setting, every smaller width transitions to the largest width in its range, i.e., $w \!\rightarrow\! w_{\max},\; \forall\, w \in \mathcal{W}\!\setminus\!\{w_{\max}\}$.
The total number of configurations is $|\mathcal{C}| = |\mathcal{W}| + (|\mathcal{W}|-1) \cdot K$, where $K$ is the number of candidate split layers. Reported values correspond to $K\!=\!1$.}
\label{tab:elastic-ranges}
\small
\begin{tabular}{lccccc}
\toprule
Model & $\mathcal{W}$ & Hybrid & $L$ & Params Range (M) & $|\mathcal{C}|$ \\
\midrule
EV-JEPA-B & $\mathcal{W}_1:\{192, 384, 768, 1024\}$ & $w \!\rightarrow\! 1024$ & 12 & 5.6--152.6 & 7 \\
EV-JEPA-L & $\mathcal{W}_2:\{192, 384, 768, 1024, 1280\}$ &  $w \!\rightarrow\! 1280$ & 12 & 5.6--238.0 & 9 \\
EV-JEPA-H & $\mathcal{W}_3:\{192, 384, 768, 1024, 1280, 1408\}$ &  $w \!\rightarrow\! 1408$ & 14 & 6.5--335.5 & 11 \\
\bottomrule
\end{tabular}
\end{table}

\section{Experiments}
\label{s:experiments}

\subsection{Experimental Setup}
\label{sec:experimental-setup}

\paragraph{Data and teacher.}
We pretrain all elastic backbones using the masked latent prediction objective described in Section~\ref{ss:vjepa_distillation}.
The pretraining set is the union of unlabeled videos proposed in V-JEPA~\cite{bardes2024vjepa}, including Something-Something~V2~(SSv2)~\cite{goyal2017something} Kinetics-710~(K710)~\cite{kay2017kinetics}, and HowTo100M~\cite{miech2019howto100m}.
Each input clip contains 16 frames sampled at 4 FPS and resized to $256 \times 256$.
Videos are tokenized with $16 \times 16 \times 2$ tubelets, so each token spans a $16 \times 16$ spatial patch over two consecutive frames.
Masking follows the long-range spatiotemporal strategy of V-JEPA~\cite{bardes2024vjepa}: we sample two spatial block masks with scale up to $70\%$ and extend them across the full temporal duration of the clip.
Target representations are produced by a frozen V-JEPA 2 ViT-G teacher that receives the same input clip as the student but unmasked. \\
\noindent\textbf{Model configurations.}
All elastic models instantiate the same elastic transformer architecture of Section~\ref{ss:elastic_arch}, with a fixed attention-head dimension of~64 and an MLP expansion ratio of~$4{\times}$.
As summarized in Table~\ref{tab:elastic-ranges}, we train three EV-JEPA model families.
EV-JEPA-B is our base model, with set widths $\mathcal{W}_1$ ranging from 192 to 1024, 12 layers, and 5.6M--152.6M parameters.
EV-JEPA-L uses the same 12-layer depth using the set of widths $\mathcal{W}_2$  with maximum width of 1280, covering approximately 5.6M--238M parameters.
EV-JEPA-H is our largest configuration, with set of widths $\mathcal{W}_3$ ranging from 192 to 1408, 14 layers, and approximately 6.5M--335.5M parameters.
LayerNorm parameters are shared across widths, and no width-specific normalization is used.
Hybrid configurations employ the shared elastic projection~$\mathbf{P}$, sliced according to the source and destination widths.
Unless otherwise noted, hybrid execution uses the middle split layer, i.e., layer~6 for 12-layer models.
In our default many-to-one setting (i.e., the hybrid setting only supports going from smaller widths to $w_{max}$) with a single fixed split point, the configuration set contains $|\mathcal{W}|$ standalone configurations and $|\mathcal{W}|-1$ hybrid paths into the largest width (i.e. $|\mathcal{C}| = |\mathcal{W}| + (|\mathcal{W}|-1)$).
More generally, with $K$ candidate split points, the number of configurations is $|\mathcal{C}| = |\mathcal{W}| + (|\mathcal{W}|-1)K$. \\
\noindent\textbf{Training strategies.}
We compare four training strategies.
The \emph{standalone} baseline trains a separate fixed-width model for each width; each standalone model is trained for the same number of epochs and with the same per-model compute capacity as the elastic models.
The three elastic variants, \emph{MatFormer-style EV-JEPA}, \emph{US-Net-style EV-JEPA}, and our proposed \emph{T3 EV-JEPA}, share the same architecture, teacher, data, masking, optimizer, and configuration set.
They differ only in how configurations are selected during pretraining, as formalized in Section~\ref{ss:pts}. \\
\noindent\textbf{Optimization.}
All EV-JEPA models are pretrained with 300 epochs and AdamW ($\beta_1\!=\!0.9$, $\beta_2\!=\!0.999$) in
  \texttt{bf16} precision on 128~GPUs (16~nodes $\times$ 8), with a constant weight decay of
  $0.04$. The learning rate follows a three-phase schedule: a linear warmup from
  $1\times10^{-4}$ to the peak, a constant phase at the peak, and a half-period cosine anneal toward
  a final learning rate of $5\times10^{-5}$. Following V-JEPA, the cosine period is stretched by a factor of
  $1.25$ relative to the training length, so training stops before the anneal completes and
  the final learning rate remains above. Schedule lengths and peak learning rates
  differ by student family. See Table~\ref{tab:optim} for the detailed training configurations for each student family.
  Inputs are 16-frame clips at $256\times256$ sampled at 4~fps, drawn from a mixture of Kinetics-710, Something-Something~v2 and HowTo100M with sampling weights $0.335/0.100/0.565$.

  \begin{table}[h]
  \centering
  \small
  \begin{tabular}{lcccccr}
  \toprule
  Student & Layers & $w_{max}$ & Warm/Const/Decay & Peak LR & Clip & Batch/GPU \\
  \midrule
  EV-JEPA-B/L  & 12 & 1024/1280 & 40\,/\,200\,/\,60  & $5.25\times10^{-4}$ & $0.025$  & 24 \\
  EV-JEPA-H & 14 & 1408 & 40\,/\,200\,/\,60 & $5.25\times10^{-4}$ & $0.025$ & 24 \\
  CNN     & 20 & 384  & 20\,/\,80\,/\,200  & $3\times10^{-3}$    & $0.025$ & 12 \\
  \bottomrule
  \end{tabular}
  \caption{Optimization schedules by student family. All runs decay to a final learning rate
  of $5\times10^{-5}$.}
  \label{tab:optim}
  \end{table}

\subsection{Elastic Training and Range Scaling on SSv2}
\label{ss:ssv2}
%
\begin{figure}[t]
    \centering
    \includegraphics[width=\linewidth]{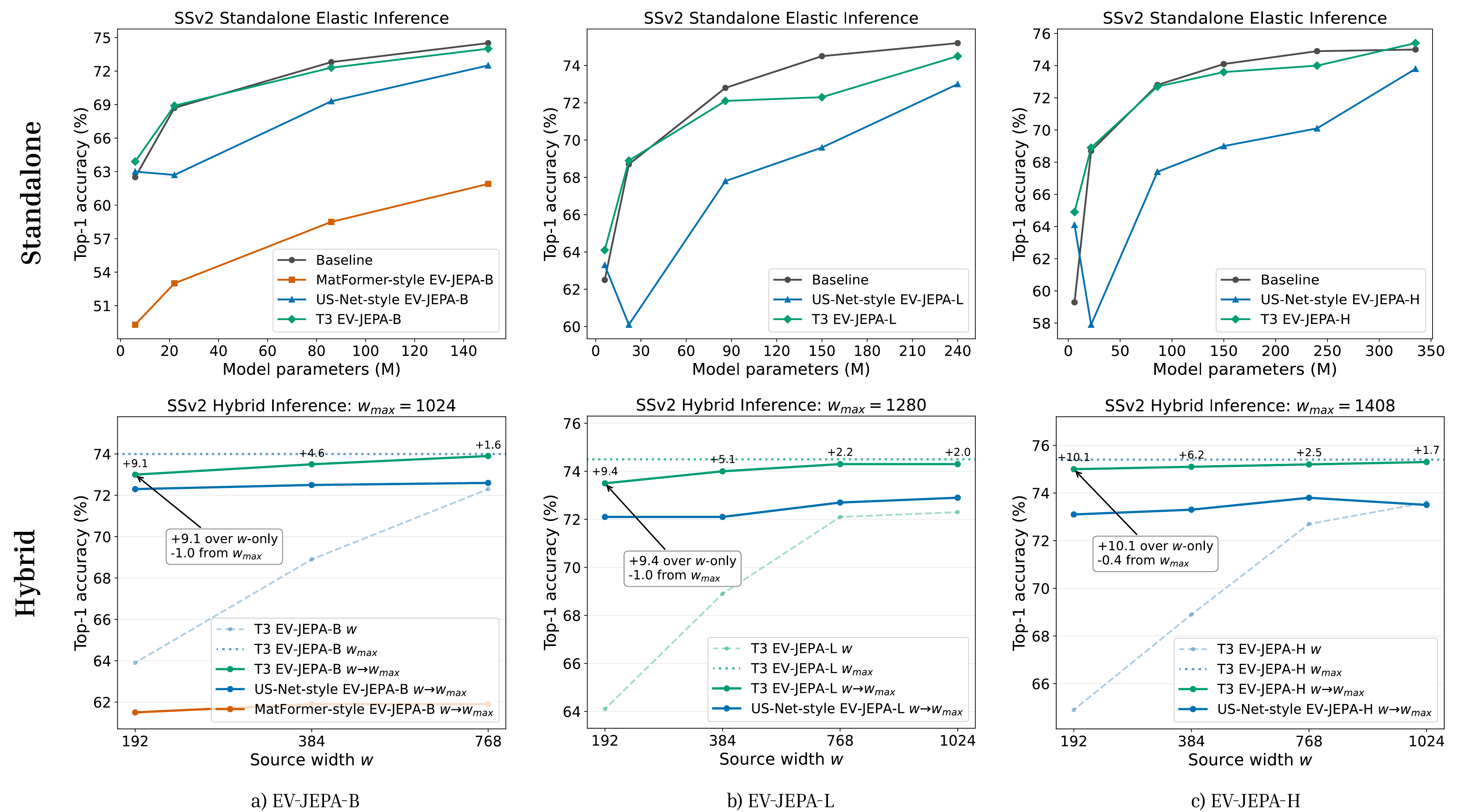}
    \caption{
    \textbf{SSv2 standalone and hybrid inference across increasing elastic ranges.} Columns correspond to a) EV-JEPA-B ($\mathcal{W}_1$), b) EV-JEPA-L ($\mathcal{W}_2$), and c) EV-JEPA-H ($\mathcal{W}_3$). The top row reports standalone elastic inference as a function of model parameters.
    The bottom row reports many-to-one hybrid execution $w \rightarrow w_{\max}$, where the first half of the encoder runs at source width $w$, and the remaining layers run at $w_{max}$.  In the bottom row, \emph{T3 EV-JEPA $w$} and \emph{T3 EV-JEPA $w_{\max}$} are shown as $w$-only and full-width T3 references, respectively, to measure hybrid recovery.
    }
    \label{fig:ssv2_ranges}
\end{figure}

We begin our evaluation on SSv2, a benchmark whose temporal sensitivity requires combining spatial recognition with fine-grained motion reasoning, making it a rigorous testbed for elastic representations that must remain useful across widely different capacities. Three questions guide our analysis. First, under a controlled setting in which all methods share the configurations, how does T3 compare to MatFormer- and US-Net-style training? Second, does this behavior remain stable as the elasticity scales from EV-JEPA-B ($\mathcal{W}_1$) to EV-JEPA-L ($\mathcal{W}_2$) and EV-JEPA-H ($\mathcal{W}_3$)? Third, does the hybrid inference help smaller models recover performance when transitioning to $w_{max}$? 

Figure~\ref{fig:ssv2_ranges}(a) \textbf{\underline{Standalone}} row reports the controlled comparison on EV-JEPA-B for the set of widths $\mathcal{W}_1$ and addresses our first question. Notably, MatFormer-style training trails the other regimes by a wide margin, suggesting that sampling a single configuration and having isolated parameter updates in each step produces unstable performance across configurations. US-Net-style sandwich training mitigates this by supervising the boundary widths at every iteration, ensuring that all shared parameters receive a gradient signal at each step. However, it comes at the cost of configurations between the two endpoints degrading. T3 avoids both failure modes. By coupling a small path ($f^{w_s}$), a larger path ($f^{w_l}$)\footnote{Note, for this specific experiment settings, $w_l = w_{max}$.}, and its hybrid transition ($f^{(w_s, w_l, k)}$) within the same iteration, every parameter is supervised at every step, and intermediate widths are explicitly drawn into the gradient. As a result, T3 tracks the independently trained baseline references most closely across the range.

Having established US-Net as the strongest competing baseline, we use it as the main benchmark in Figures~\ref{fig:ssv2_ranges} b) and~\ref{fig:ssv2_ranges} c), which extend the elastic range to $\mathcal{W}_2$ with EV-JEPA-L and $\mathcal{W}_3$ with EV-JEPA-H. The trend is clear, T3 stays close to the references throughout, whereas US-Net progressively worsens at intermediate widths, with the gap reaching up to $11$ accuracy points in EV-JEPA-H. Endpoint supervision alone is therefore increasingly insufficient as the range grows. T3, in contrast, drops the fixed-boundary anchoring and provides a stable optimization signal across the entire range. 
%


The \textbf{\underline{Hybrid}} row of Figure~\ref{fig:ssv2_ranges} addresses our third question. Because hybrid accuracy is upper-bounded by the destination width $w_{\max}$, two criteria are relevant: whether T3 outperforms US-Net-style training, and whether a low-width model can be compatible with a high-capacity continuation to recover the performance gap with the $w_{max}$ standalone model.  To make this recovery explicit, we plot two T3 references alongside the hybrid curve: the standalone $w$-only execution as a lower bound and the standalone $w_{\max}$ execution as an upper bound. The vertical distance from the $w$-only curve to the hybrid curve thus quantifies the accuracy gained by continuing computation at $w_{\max}$, while the distance from the hybrid curve to the $w_{\max}$ reference captures the residual loss relative to executing the $w_{max}$ model end-to-end. Across all three ranges, T3 hybrid execution recovers nearly $w_{max}$-level accuracy even with source widths as small as $w=192$. Particularly, EV-JEPA-H ($w=192$) model gains over $10$ accuracy points when continued to $w_{\max}$ and falls within $1$ point of the standalone $w_{max}$ model, with analogous trends in EV-JEPA-B and EV-JEPA-L. Cross-width compatibility is therefore preserved across different ranges, validating T3 as a principled training regime for distributed inference.

\subsection{Evaluation on Recognition (K400) and Action Anticipation (EK100)}
\label{sec:k400_ek100}

\begin{figure}[t]
    \centering
    \includegraphics[width=\linewidth]{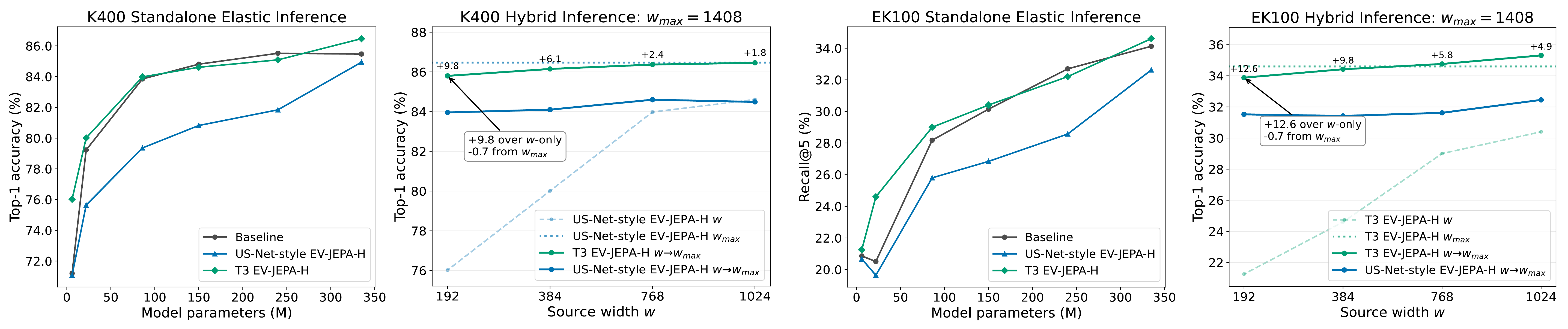}
    \caption{\textbf{Standalone and hybrid inference on K400 for action recognition and EK100 for action anticipation.} \emph{Standalone} panels (1st, 3rd) report elastic inference as a function of model parameters. \emph{Hybrid} panels (2nd, 4th) report hybrid execution $w\!\rightarrow\!w_{\max}$, where the first half of the encoder runs at source width $w$, and the remaining layers continue at $w_{\max}$. 
    }
    \label{fig:k400_ek100}
\end{figure}

Having analyzed elastic and hybrid behavior on SSv2, we now ask whether similar conclusions transfer to settings with a different discriminative structure. K400 is appearance- and scene-biased, stressing spatial--semantic capacity, while EK100 moves from recognition to \emph{action anticipation}. For action anticipation, we use frozen pretrained elastic predictor (following equation \ref{eq:predictor-width} convention) for evaluation. Following the analysis in Section~\ref{ss:ssv2}, we retain US-Net as the strongest competing elastic baseline, and we report all results on the full-range EV-JEPA-H model ($\mathcal{W}_3$, $w_{\max}=1408$). K400 is evaluated with Top-1 accuracy and EK100 with mean Top-5 recall (recall@5), the standard action-anticipation metric~\cite{furnari2020rolling,diko2024semantically}.

\paragraph{Standalone.}
The standalone panels of Fig.~\ref{fig:k400_ek100} (1st, 3rd) show that on both K400 and EK100, T3 tracks the independently trained references across the full width range, whereas US-Net degrades at intermediate widths, with the gap reaching up to ~5 accuracy points on K400 and ~5 recall@5 points on EK100.

\paragraph{Hybrid.}
The hybrid panels (2nd, 4th) report many-to-one execution $w\!\rightarrow\!w_{\max}$, with the T3 $w$-only and $w_{\max}$ standalone curves plotted as lower and upper bounds. On both benchmarks, continuing a small-width encoder at $w_{\max}$ recovers most of the gap to the standalone $w_{\max}$ model. Even at source width $w=192$, hybrid execution gains +9.8 Top-1 points on K400 and +12.6 recall@5 points on EK100, falling within 1 point(s) of the full-width reference, while consistently outperforming US-Net-style hybrid execution. Cross-width compatibility, therefore, holds for appearance-biased recognition and for egocentric anticipation alike, supporting T3 as a general training regime for distributed inference rather than one tuned to a single benchmark.

\subsection{Any-to-Any Hybrid Execution}
\label{sec:any2any}

\begin{figure}[!t]
    \centering
    \includegraphics[width=1.0\linewidth]{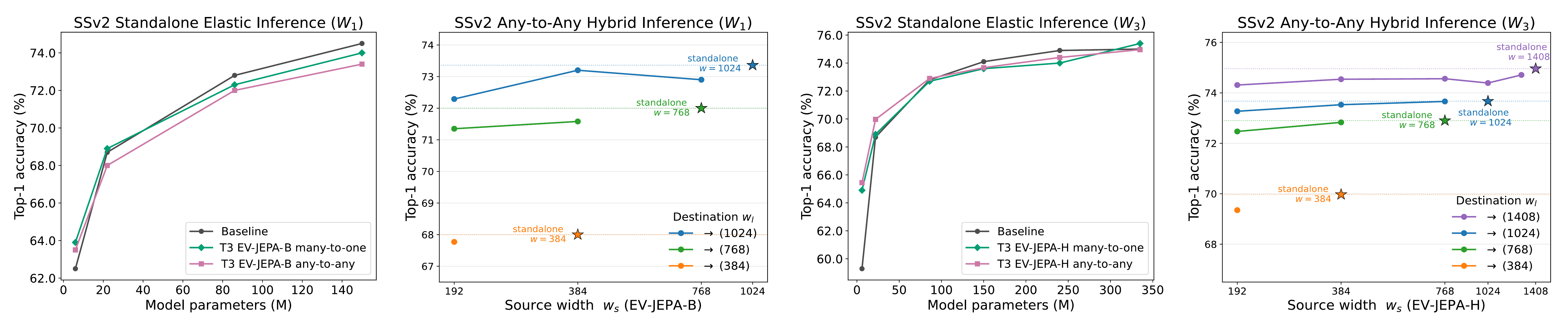}
    \caption{\textbf{SSv2 any-to-any hybrid inference.} In the any-to-any setting, a model executing at any source width $w_s$ can upscale at \emph{any} larger width $w_l > w_s$. On the \textbf{\textit{Hybrid Inference}} plots, each curve fixes a destination $w_l$ and sweeps the source $w_s$. The star and dotted line mark the standalone accuracy of that destination width. Across all pairs, hybrid execution recovers close to each destination's standalone accuracy, even at $w_s{=}192$, which is within $\sim$1 point of the standalone largest width. Accuracy is therefore governed by the destination (continuation) width, confirming that a small device can pair with any larger device at minimal loss.}
    \label{fig:any2any}
\end{figure}

The hybrid experiments so far adopt the many-to-one setting ($w\!\rightarrow\!w_{\max}$), our default throughout the paper. Here we verify the more general claim that computation can transition from \emph{any} source width to \emph{any} larger destination width. Because the any-to-any configuration set is more expensive to train than the many-to-one default, we conduct this study on the smallest and largest ranges, $\mathcal{W}_1=\{192,384,768,1024\}$ and $\mathcal{W}_3=\{192,384,768,1024, 1280, 1408\}$, with a single mid-network split ($k=6$ for the 12-layer model $\mathcal{W}_1$ and $k=7$ for the 14-layer model $\mathcal{W}_3$). Figure~\ref{fig:any2any} reports SSv2 Top-1 for every valid source--destination pair $(w_s, w_l)$ with $w_s < w_l$. Note that the upscaling projection layer $P_{w_s \rightarrow w_l} $ is the same for each configuration, as in the many-to-one approach.

Two observations emerge. First, the any-to-any T3 model retains strong standalone quality across the full width range, its standalone accuracy closely tracks both the independently trained baselines and the many-to-one T3 elastic network (Fig.~\ref{fig:any2any} \textbf{\textit{left}}), incurring only a small trade-off of less than $1$ point relative to the many-to-one model (in some configurations the any-to-any performs better than the many-to-one) as the price of supporting the larger any-to-any configuration space. Second, hybrid accuracy is governed primarily by the destination width $w_l$ (similar to the many-to-one T3 model). For $w_l{=}1024$, all sources fall within $0.9$ points of one another and within $1.1$ points of the standalone $w{=}1024$ reference ($73.4$), including the smallest source $w_s{=}192$ at $72.3$. On the other hand, for $w_l{=}1408$ ($\mathcal{W}_3$), all sources fall within $0.3$ points of one another and within $0.65$ points of the standalone $w{=}1408$ reference ($74.96$). Together, these confirm that a small originating device can hand off to an arbitrarily larger continuation device and recover near-destination accuracy, extending the many-to-one result of Section~\ref{ss:ssv2} to the full any-to-any setting at negligible standalone cost in some configuration.

\subsection{Variable Exit Points and Elastic Token Compression}
\label{ss:compression}

\begin{figure}[!t]
    \centering
    \includegraphics[width=\linewidth]{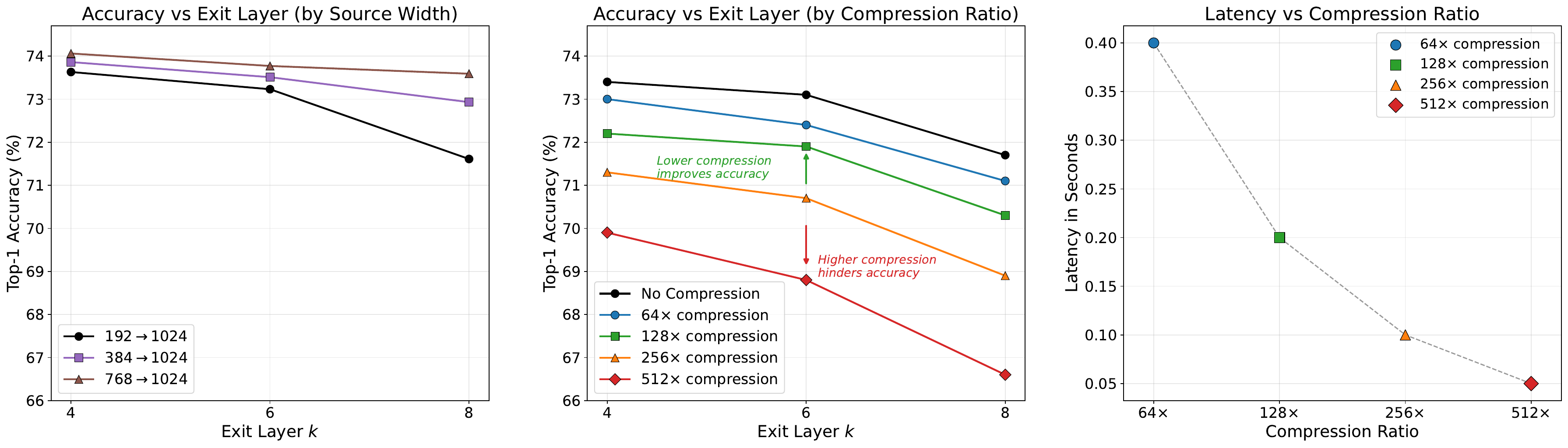}
        \caption{\textbf{Exit point, source width, and elastic token compression
        trade-offs.} \textbf{(left)} SSv2 Top-1 for hybrid execution as a function of the
        exit layer $k$, for source widths $w_\in\{192,384,768\}$ and target width $w_l=1024$.
        \textbf{(middle)} SSv2 Top-1 versus $k$ for the single path
        $192{\rightarrow}1024$ under different elastic compression ratios. \textbf{(right)}
        Transmission latency at each compression ratio over a theoretical 1 Mbps BLE
        link, computed as the transmitted payload divided by link bandwidth.
        }
    \label{fig:tradeoffs}
\end{figure}

\begin{figure}[!ht]
    \centering
    \includegraphics[width=\linewidth]{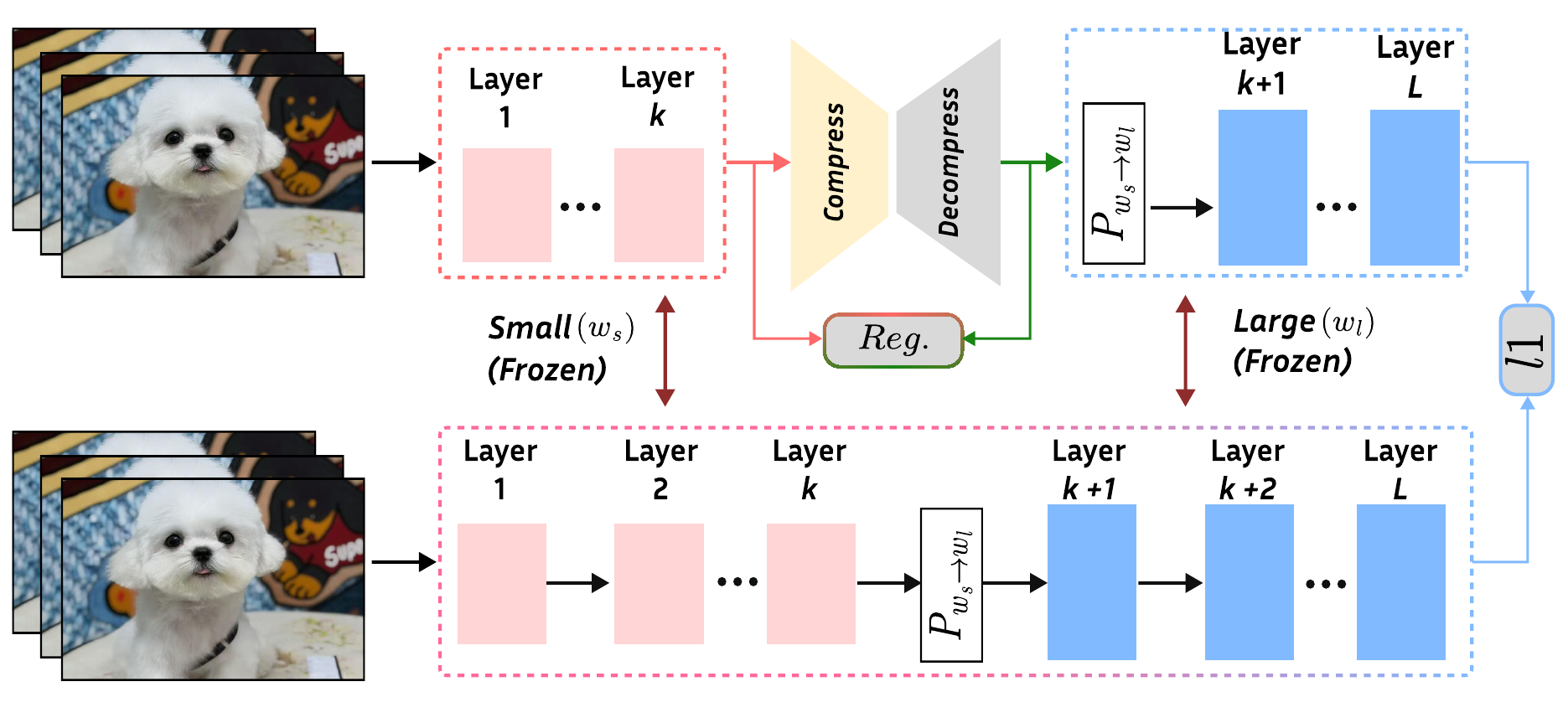}
    \caption{%
       \textbf{Elastic token compression for hybrid execution.} A learned compressor is inserted at the exit point $k$ of the source device. The 'compress module' compresses the intermediate representation before transmission, reducing
 the number of tokens; the 'decompress module' reconstructs the token grid on the target device, after which
the shared elastic projection $P_{w_s \to w_l}$ upscales to $w_l$, and the remaining $L-k$ layers execute. During training, all modules except the compress–decompress pair remain frozen.}
    \label{fig:compression_arch}
\end{figure}

\paragraph{Variable splitting layer \textit{k}.}
%
The hybrid experiments so far have relied on a fixed exit point $k$, placing the transition at the middle of the network. The architecture, however, does not constrain us to a single split layer: because the same projection $P$ is reused for every $k$ (Section~\ref{ss:pts}), supporting multiple exit points adds no parameters. We therefore conduct a variable-$k$ experiment on our $\mathcal{W}_1$ configurations, with widths ranging from $192$ to $1024$ and exit points $k \in \{4, 6, 8\}$ in a $12$-layer network. Figure~\ref{fig:tradeoffs}~(\emph{left}) reports SSv2 Top-1 as a function of the exit layer for three hybrid paths $w \rightarrow 1024$ with $w \in \{192, 384, 768\}$; each curve fixes the source width and sweeps $k$. Two effects are visible. First, accuracy decreases monotonically with $k$ for every source width: a later exit leaves fewer layers to the larger continuation width $w_l$, whereas an earlier handoff leaves more of the network at $w_l$ and recovers more of its standalone accuracy. Second, how much a late exit costs is governed by the source width. At the earliest exit ($k\!=\!4$) the three paths are nearly indistinguishable, within $0.43$ points of one another, so even the narrowest $w\!=\!192$ source reaches essentially the accuracy of a $4\times$ wider one; by $k\!=\!8$ the curves fan out and the $w\!=\!192$ path degrades the most, falling $2.0$ points below $w\!=\!768$. On-device capacity and on-device depth are therefore substitutable: the earlier the handoff, the less the source width matters. This exposes the exit point as a deployment knob, while the policy that selects an operating point is application-specific and outside our scope. The compressor is trained in a separate post-training stage with all remaining network elements frozen, using an $\ell_1$ reconstruction loss shown in figure~\ref{fig:compression_arch}. 

\paragraph{Elastic token compression.}
Hybrid execution assumes the transmission of an intermediate representation from the source device to the continuation device, where the communication of a large amount of data can incur high latency and high power consumption. To reduce transmitted data, we attach a learned compressor at the exit of the source device, realized as a single encoder-decoder 3D convolution (more details in the appendix). The encoder runs as one additional layer and the decoder reconstructs the representation on the destination device before the remaining $L-k$ layers execute. The compressor reduces the number of transmitted tokens. Following the nested weight-sharing scheme of Section~\ref{ss:elastic_arch}, compression is made elastic by slicing the convolution kernel along its spatiotemporal dimension,
\begin{equation}
\label{eq:kernel}
K^{(\kappa)} = K\big[\,:,\ :,\ 1{:}\kappa_t,\ 1{:}\kappa_h,\ 1{:}\kappa_w\,\big],
\end{equation}
where the input and output channel dimensions are fixed at $C_\text{in}=C_\text{out}=w_s$, and the sliced $(\kappa_t,\kappa_h,\kappa_w)$ and strides set the spatiotemporal downsampling factor and hence the number of retained tokens. A single set of kernel weights thus spans the entire range of compression ratios, so one on-device layer can adjust compression elastically at inference time in the same way the backbone adjusts width.

\paragraph{Compressor configuration.} 
We instantiate the compressor on EV-JEPA-B ($\mathcal{W}_1=\{192,384,768,1024\}$, $L=12$) for the hybrid configuration $f_\theta^{(w_s,w_l,k)}$ with source width $w_s=192$, destination width $w_l=w_{\max}=1024$ at a given split-layer $k$, so the compressor sits exactly where the source device hands off to the continuation device. Up to layer $k$ the encoder $E_\theta$ runs at $w_s$, so the quantity to be transmitted is the intermediate representation $h_k^{(w_s)}\in\mathbb{R}^{N\times w_s}$. For a $16$-frame $256\times256$ clip tokenized with $16\times16\times2$ tubelets, this is $N=T\times H\times W=8\times16\times16=2048$ tokens of width $w_s=192$. The \emph{compress} module reshapes these $N$ tokens back into a $w_s\times T\times H\times W$ volume and applies a single strided $3$D convolution with the sliced kernel $K^{(\kappa)}$ of Eq.~\ref{eq:kernel}, followed by GroupNorm ($8$ groups) and GELU. The sliced extent $(\kappa_t,\kappa_h,\kappa_w)$, together with the matching stride $(\sigma_t,\sigma_h,\sigma_w)$ and per-axis padding $p=\lfloor(\kappa-\sigma)/2\rfloor$, fixes the retained grid $N'=T'\times H'\times W'$ and hence the token-reduction factor $\tau=N/N'$, while the channel dimension/width is kept to $C_{\mathrm{out}}=w_s=192$. The $N'$ tokens of width $w_s$ are what crosses the link, giving the payload $P_{\mathrm{comp}}$ of Eq.~\ref{eq:comp}. The \emph{decompress} module mirrors this on the continuation device: a $3\times3\times3$ convolution with GroupNorm and GELU, a strided $3$D transposed convolution with GroupNorm and GELU that restores the original $T\times H\times W$ grid at width $w_s$, and a LayerNorm over the channel dimension. The reconstructed tokens $\hat{h}_k^{(w_s)}$ then enter the same shared elastic projection $P_{w_s\to w_l}=P[1{:}w_l,\,1{:}w_s]$ of Sec.~\ref{ss:elastic_arch} (here $P_{192\to1024}$), after which the remaining $L-k$ layers execute at $w_l$, exactly as in uncompressed hybrid execution (Fig.~\ref{fig:compression_arch}). Elasticity follows the same nested weight-sharing policy as the backbone: the strided convolution and the transposed convolution each store a single $6\times6\times6$ kernel $K$, and every token-reduction factor $\tau$ selects the leading sub-kernel $K^{(\kappa(\tau))}$ via Eq.~\ref{eq:kernel}. One set of kernel weights therefore spans the whole range, so $\tau$ becomes an inference-time knob in the same sense as the standalone width $w$, the hybrid pair $(w_s,w_l)$, and the split layer $k$. Table~\ref{tab:ratios} lists $(\kappa,\sigma,p)$, the retained grid $N'$, and the corresponding payload ratio $\rho=8\tau$ for each configuration.


\begin{table}[t] 
\centering 
\caption{Per-ratio configurations of the elastic encoder--decoder for an input grid of $8\times16\times16=2048$ tokens. Kernel, stride, and padding are given as (time, height, width). $\tau=N/N'$ is the token-reduction factor set by the sliced kernel (Eq.~\ref{eq:kernel}), and $\rho$ is the resulting payload compression ratio relative to raw RGB (Eq.~\ref{eq:comp}), which is the quantity reported in Fig.~\ref{fig:tradeoffs}.} 
\label{tab:ratios} 
\begin{tabular}{lccccc} 
\toprule $\tau$ & Kernel & Stride & Padding & Latent grid ($T'\!\times\!H'\!\times\!W'$) / tokens & $\rho$ \\ \midrule 
$8\times$ & $(4,4,4)$ & $(2,2,2)$ & $(1,1,1)$ & $4\times8\times8$ / $256$ & $64\times$ \\ 
$16\times$ & $(6,4,4)$ & $(4,2,2)$ & $(1,1,1)$ & $2\times8\times8$ / $128$ & $128\times$ \\ 
$32\times$ & $(4,6,6)$ & $(2,4,4)$ & $(1,1,1)$ & $4\times4\times4$ / $64$ & $256\times$ \\ 
$64\times$ & $(6,6,6)$ & $(4,4,4)$ & $(1,1,1)$ & $2\times4\times4$ / $32$ & $512\times$ \\ 
\bottomrule 
\end{tabular} 
\end{table}

\paragraph{Compressor training.} The compressor is trained in a second, post-training stage on top of a T3-pretrained EV-JEPA-B (stage 1), with every other module frozen, so the elastic encoder behaves exactly as during pretraining and the reported numbers isolate the cost of compression alone. Writing $\theta$ for the frozen pretrained parameters, the reference path is the uncompressed hybrid configuration $f_\theta^{(w_s,w_l,k)}$ and the trained path is the same configuration with the compress--decompress pair inserted at layer $k$, denoted $f_\theta^{(w_s,w_l,k,\tau)}$ at token-reduction factor $\tau$; only the compressor parameters receive gradients. Two things differ from Sec.~\ref{ss:vjepa_distillation}. First, this stage operates on unmasked clips: there is no masked index set $\mathcal{M}$, no mask token $\mathbf{m}$, and no predictor $G_\theta$, so supervision is applied at the encoder output over \emph{all} $N$ tokens rather than over $\mathcal{M}$ alone. Second, the objective is the $\ell_1$ regression of Eq.~\ref{eq:l1} \emph{regularized} by a reconstruction term at the split layer, which anchors the transmitted representation itself rather than only its downstream effect. Concretely, let $h_k^{(w_s)}\in\mathbb{R}^{N\times w_s}$ denote the source tokens at split layer $k$ and $\hat{h}_{k\tau}^{(w_s)}\in\mathbb{R}^{N\times w_s}$ denote their reconstruction after compress--decompress at factor $\tau$, and let $z^\tau$ and $z$ denote the layer-normalized final encoder features of the compressed and uncompressed (frozen model from pretraining or stage 1) paths on an unmasked clip $x$. For a single $\tau$, 
\begin{equation} \label{eq:compression}
\mathcal{L}_{\mathrm{comp}}(\tau)= 
\underbrace{\frac{1}{N}\sum_{i=1}^{N} \big\|\,z_i^\tau-z_i\,\big\|_1}_{\textbf{L1 Term (Eq.~\ref{eq:l1})}} + \underbrace{\frac{1}{N}\sum_{i=1}^{N}\big\|\, h_{k,i}^{(w_s)} - \hat{h}_{k\tau,i}^{(w_s)}\,\big\|_1}_{\text{Regularization Term}},
\end{equation}
and the stage optimizes the mean over the set of token-reduction factors $\mathcal{R}=\{8,16,32,64\}$ (equivalently $\rho\in\{64\times,128\times,256\times,512\times\}$): 
\begin{equation} \label{eq:comp_full}
\mathcal{L}_{\mathrm{comp}}=\frac{1}{|\mathcal{R}|} \sum_{\tau\in\mathcal{R}}\mathcal{L}_{\mathrm{comp}}(\tau). 
\end{equation}  
At every iteration the trained path is run once at each $\tau\in\mathcal{R}$ and the gradients are summed before a single optimizer step. This mirrors the T3 principle of Sec.~\ref{ss:pts} one level down: rather than sampling one ratio per step, all ratios are supervised jointly, so the shared kernel $K$ is optimized for every compression level at once and no ratio is left to compete for a sampling slot. Optimization: $100$ epochs of $300$ iterations with a global batch of $4096$ clips drawn from the same pretraining corpus (K710, SSv2, and HowTo100M, sampling weights $0.335/0.100/0.565$) in \texttt{bf16}, AdamW with weight decay $0.04$, a $10$-epoch linear warmup from $10^{-4}$ to $5.25\times10^{-4}$ and cosine decay to $5\times10^{-5}$.

\paragraph{Accuracy under compression.} Figure~\ref{fig:tradeoffs}~(\emph{middle}) isolates a single hybrid path, $w_s\!=\!192 \rightarrow w_l\!=\!1024$, and sweeps the same exit points $k \in \{4, 6, 8\}$ at four compression ratios --- $64\times$, $128\times$, $256\times$, and $512\times$ --- with the uncompressed hybrid path plotted as the \emph{No-Compression} reference. Two observations follow. First, the ordering with respect to $k$ is preserved at every ratio: all curves decrease with a later exit and none cross, so the exit point and the compression ratio behave as two independent knobs rather than interacting ones. Second, accuracy decreases monotonically with the compression ratio, and the degradation is graceful at the low end: at $64\times$ the compressed path tracks the No-Compression curve within $0.7$ Top-1 points, and only at $512\times$ does the penalty grow to $5.1$ points.

\paragraph{Payload and trade-off.}
We define compression relative to transmitting the input as raw $8$-bit RGB, the worst case for an always-on sensor without a hardware video codec. For an intermediate representation of $N$ tokens reduced to $N'$ tokens of dimension $w_s$, the transmitted payload is $P_\text{comp}=N'\cdot w_s\cdot b$ bits, and the compression ratio is
\begin{equation}
\label{eq:comp}
\rho = \frac{P_\text{RGB}}{P_\text{comp}}, \qquad P_\text{RGB} = F\cdot H\cdot W\cdot 3 \cdot 8,
\end{equation}
with $F$, $H$, $W$ the clip length and frame resolution and $b=8$ the bits per transmitted element. The elastic token reduction reaches up to $512\times$ payload reduction relative to raw RGB. All reported accuracies are end-to-end, measured after the decoder reconstructs the token grid and the continuation layers run, so they reflect the true cost of compressing the transmitted representation. Together, the exit point and the compression ratio form two orthogonal knobs that expose a controllable trade-off among on-device computation, transmission latency, and accuracy. Regarding the compression network, it is trained in a separate post-training stage with all the remaining network elements frozen using an L1 reconstuction loss.

\subsection{Comparison With SOTA}
\label{ss:sota}

\begin{table}[h]
    \centering
    \small
    \setlength{\tabcolsep}{14pt}          
    \renewcommand{\arraystretch}{0.95}    
    \caption{Comparison with state-of-the-art models for action recognition on SSv2 and K400. \best{Bold} marks the best result per column,
    \second{underline} the second best. Avg.\ is the mean of SSv2 and K400.}
    \label{tab:main}
    \begin{tabular}{l r S[table-format=2.1] S[table-format=2.1] S[table-format=2.1] }
    \toprule
    & & \multicolumn{1}{c}{\gr{Motion}} & \multicolumn{1}{c}{\gr{Appearance}} & \\
    \cmidrule(lr){3-3}\cmidrule(lr){4-4}
    Method & Param. & {SSv2} & {K400} & {Avg.} \\
    \midrule
    \multicolumn{5}{l}{\gr{Image encoders Evaluated Using Same Protocol}}\\
    DINOv2~\cite{oquab2023dinov2} & 1.1B & 50.7 & 83.6 & 67.2 \\
    PE\textsubscript{core}G~\cite{bolya2026perception}          & 1.9B & 55.4 & \second{88.5} & 72.0 \\
    SigLIP2~\cite{tschannen2025siglip}                           & 1.2B & 49.9 & 87.3 & 68.6 \\
    \addlinespace[1pt]
    \multicolumn{5}{l}{\gr{Video encoders Evaluated Using Same Protocol}}\\
    VideoMAEv2~\cite{wang2023videomae}                        & 1.0B & 56.1 & 82.8 & 69.5 \\
    InternVideo2-1B~\cite{wang2024internvideo2}                   & 1.0B & 67.3 & 87.9 & 77.6 \\
    VideoPrism~\cite{zhao2024videoprism}                        & 1.0B & 68.5 & 87.6 & 78.0 \\
    InternVideo2\textsubscript{s2}-1B~\cite{wang2024internvideo2} & 1.0B & 69.7 & \best{89.4} & 79.6 \\
    V-JEPA 2 ViT-L~\cite{assran2025vjepa}                    & 300M & 73.7 & 85.1 & 79.4\\
    V-JEPA 2 ViT-H~\cite{assran2025vjepa}                    & 600M & 74.0 & 85.3 & \second{79.7}\\
    V-JEPA 2 ViT-G~\cite{assran2025vjepa}                    & 1.0B & \second{75.3} & 86.6 & \best{81.0}\\
    \midrule
    \multicolumn{5}{l}{\gr{Distilled from V-JEPA 2 ViT-G}}\\
    \textit{Non-Elastic} Baseline ($w{=}1408$)   & 335M & 75.0 & 85.5 & 80.3\\
    \rowcolor{ourrow}
    \textbf{EV-JEPA-H} ($w{=}1408$)   & 335M & \best{75.4} & 86.5 & \best{81.0}\\
    \bottomrule
    \end{tabular}
\end{table}

\begin{table}[t]
    \centering
    \small
    \setlength{\tabcolsep}{10pt}
    \renewcommand{\arraystretch}{0.95}
    \caption{Action anticipation on EK100. \best{Bold} marks the best result per
    column. Frozen-backbone rows attach an
    anticipation head to a frozen encoder.}
    \label{tab:anticipation}
        \begin{tabular}{l r S[table-format=2.1] S[table-format=2.1] S[table-format=2.1]}
        \toprule
        & & \multicolumn{3}{c}{\gr{Action anticipation (EK100)}} \\
        \cmidrule(lr){3-5}
        Method & Param. & {Verb} & {Noun} & {Action} \\
        \midrule
        InAViT \cite{roy2024inavit}          & 160M & 51.9 & 52.0 & 25.8 \\
        Video-LLaMA \cite{zhang2023video} & 7.0B & 52.9 & 52.0 & 26.0 \\
        PlausiVL \cite{mittal2024plausivl}   & 8.0B & 55.6 & 54.2 & 27.6 \\
        \addlinespace[1pt]
        \multicolumn{5}{l}{\gr{Frozen backbone}}\\
        V-JEPA 2 ViT-L                        & 300M & 57.8 & 53.8 & 32.7 \\
        V-JEPA 2 ViT-H                        & 600M & 59.2 & 54.6 & 36.5 \\
        V-JEPA 2 ViT-g                        & 1.0B & \best{61.2} & \best{55.7} & \best{38.0} \\
        \midrule
        \multicolumn{5}{l}{\gr{Frozen backbone Distilled from V-JEPA 2 ViT-g}}\\
         \textit{Non-Elastic} Baseline ($w{=}1408$)       & 335M & 58.0 & 53.2 & 34.1 \\
        \rowcolor{ourrow}
        \textbf{EV-JEPA-H} ($w{=}1408$)       & 335M & 58.2 & 53.2 & 34.6 \\
        \bottomrule
    \end{tabular}
\end{table}

Table~\ref{tab:main} compares EV-JEPA-H at full width ($w{=}1408$) against
state-of-the-art image and video foundation models on SSv2 (motion) and K400
(appearance), all evaluated under the same attentive-probe protocol. Because
EV-JEPA-H is distilled from a frozen V-JEPA~2 ViT-g teacher, this is not a
like-for-like comparison against models trained from scratch, and we therefore
include a \emph{Non-Elastic} Baseline distilled under identical teacher, data,
masking, and compute budget. This control isolates the accuracy cost of elasticity from
the effect of distillation, which is a quantity of interest.

Read this way, three results stand out. First, elasticity is obtained at no
measurable accuracy cost. EV-JEPA-H matches the \emph{Non-Elastic} Baseline across the
board (75.4 vs.\ 75.0 on SSv2, 86.5 vs.\ 85.5 on K400, 81.0 vs.\ 80.3 average),
despite the same weights additionally serving every smaller width and every
cross-device hybrid configuration. We do not read the small margin in our favour
as evidence that elasticity \emph{improves} full-width accuracy, though it is
consistent with the multi-configuration objective of T3 acting as a regulariser.
Second, this operating point is competitive with fixed-capacity specialists:
EV-JEPA-H attains the best SSv2 accuracy (75.4), edging its own
ViT-g teacher (75.3, 1.0B) at roughly $3\times$ fewer parameters, and far
surpassing appearance-biased image encoders such as PEcore$_\text{G}$ (55.4) and
DINOv2 (50.7) on this motion-centric benchmark. On the SSv2/K400 average it
reaches 81.0, matching V-JEPA~2 ViT-g while exceeding ViT-L (79.4, 300M) and
ViT-H (79.7, 600M). Third, on the appearance-dominated K400, EV-JEPA-H (86.5)
falls within 0.1 points of ViT-g and is competitive with the strongest video
encoders, though encoders trained on larger labelled
corpora (e.g.\ InternVideo2$_\text{s2}$, 89.4) remain ahead, consistent with the
motion-versus-appearance trade-off inherent to the V-JEPA objective.


On action anticipation (EK100, Table~\ref{tab:anticipation}), EV-JEPA-H at full width
reaches 34.6 action recall@5, again matching the \emph{Non-Elastic} Baseline distilled
under identical conditions (34.1). Elasticity is therefore free on
anticipation as well as on recognition. This also clarifies how to read
the remaining gap to the frozen V-JEPA~2 ViT-H backbone (36.5): it is a capacity
effect rather than a cost of our training regime, since a fixed-width model at
the same 335M parameters and the same teacher lands below EV-JEPA-H. Relative to
the broader literature, EV-JEPA-H outperforms recent 7--8B video--language models
(PlausiVL, 27.6, and Video-LLaMA, 26.0) by roughly seven points at a fraction of
their size, and sits between the frozen ViT-L (32.7) and ViT-H (36.5) backbones
at only 335M parameters. 
Crucially, these numbers are the full-width operating point of a \emph{single elastic} model, not a separately trained specialist. The same weights that place EV-JEPA-H at or above the state of the art at $w{=}1408$ also serve every smaller width and cross-device hybrid configuration. EV-JEPA thus matches fixed-capacity state-of-the-art models at its high end while additionally offering the reconfigurable compute and collaborative execution that none of these baselines provide.

\subsection{Vision-Language Transfer with VL-JEPA}

Sections 4.2 and 4.3 probe the visual representation directly through task-specific heads. We now ask whether the same elastic encoder also transfers to vision-language understanding, where the representation must align with language rather than a discriminative head. We adopt VL-JEPA~\cite{chen2025vl}, a joint-embedding vision-language model that predicts continuous text embeddings, and replace its frozen V-JEPA 2 ViT-L vision encoder with our frozen EV-JEPA-H encoder, training only the connector and predictor. Following VL-JEPA, we report Top-1 accuracy on SSv2 open-vocabulary classification.

As shown in Table \ref{tab:vljepa}, in standalone inference, the full-width configuration ($w{=}1408$) matches and slightly exceeds the VL-JEPA ViT-L baseline, while reducing to $w{=}1024$ stays within one point at only 179M parameters, about 60\% of the ViT-L encoder. In hybrid cross-device execution, where the early layers run at a small source width and the remaining layers continue at $w_{\max}$ so that only compact intermediate features are transmitted, the $384{\to}1408$ and $192{\to}1408$ configurations reach within one point of the baseline while activating only 194M--203M parameters. A single frozen EV-JEPA encoder therefore extends both its standalone width-elasticity and its cross-device hybrid capability from recognition and anticipation to vision-language tasks, confirming that elasticity and language alignment compose without a per-budget encoder.


  \begin{table}[t]
  \centering\small
  \begin{minipage}[t]{0.44\linewidth}
  \centering
  \setlength{\tabcolsep}{10pt}
  \renewcommand{\arraystretch}{0.95}
  \caption{Vision--language transfer on SSv2 (Top-1, open-vocabulary classification) under the VL-JEPA framework.
  All elastic rows use a single frozen EV-JEPA-H encoder.}
  \label{tab:vljepa}
  \begin{tabular}{l r S[table-format=2.1]}
  \toprule
  Configuration & Param. & {SSv2} \\
  \midrule
  V-JEPA 2 ViT-L~\cite{chen2025vl} & 300M & 73.2 \\
  \midrule
  \multicolumn{3}{l}{\gr{Standalone}}\\
  \sub{$w{=}1408$}   & 335M & \best{73.6} \\
  \sub{$w{=}1024$}   & 179M & 72.2 \\
  \addlinespace[1pt]
  \multicolumn{3}{l}{\gr{Hybrid}}\\
  \sub{$384{\to}1408$} & 203M & 72.9 \\
  \sub{$192{\to}1408$} & 194M & 72.6 \\
  \bottomrule
  \end{tabular}
  \end{minipage}\hfill
  \begin{minipage}[t]{0.54\linewidth}
  \centering
  \setlength{\tabcolsep}{10pt}
  \renewcommand{\arraystretch}{0.95}
  \caption{Elasticity beyond Transformers. SSv2 Top-1 for a slimmable factorized 3D-CNN with a lightweight
  linear-attention layer. A single nested model serves all standalone widths and their cross-width hybrids.}
  \label{tab:cnn}
  \begin{tabular}{l r r S[table-format=2.1]}
  \toprule
  Configuration & Param. & GFLOPs & {SSv2} \\
  \midrule
  \multicolumn{4}{l}{\gr{Standalone}}\\
  \sub{$w{=}128$}  & 2.3M  & 11 & 56.0 \\
  \sub{$w{=}192$}  & 4.1M  & 20 & 58.6 \\
  \sub{$w{=}256$}  & 6.4M  & 30 & 60.7 \\
  \sub{$w{=}384$}  & 12.2M & 57 & \best{63.5} \\
  \addlinespace[1pt]
  \multicolumn{4}{l}{\gr{Hybrid}}\\
  \sub{$128{\to}384$} & 11.7M & 50 & 63.2 \\
  \sub{$192{\to}384$} & 11.8M & 51 & 63.2 \\
  \sub{$256{\to}384$} & 12.0M & 53 & 63.4 \\
  \bottomrule
  \end{tabular}
  \end{minipage}
  \end{table}

\subsection{Elasticity Beyond Transformers: Convolutional Backbones}
\label{sec:cnn}


The transition-aware training behind EV-JEPA is not specific to the Transformer. We distill the ViT-g teacher into a slimmable factorized 3D-CNN, augmented with a single lightweight linear-attention layer for global context. The CNN channel widths are sliced under the same weight-sharing policy (Section~\ref{ss:elastic_arch}), so smaller subnets are strict channel slices of the largest and one set of weights serves every width. We use channels-first LayerNorm so that a single set of statistics-free normalizers stays valid across widths. Trained with the same T3 regime (Section~\ref{ss:pts}), a single nested CNN scales monotonically on SSv2 (Table~\ref{tab:cnn}): from 56.0 at $w{=}128$ (2.3M parameters, 11 GFLOPs), to 63.5 at $w{=}384$ (12.2M, 57 GFLOPs). This extends EV-JEPA's elasticity to the low-cost, low-memory regime most relevant to always-on, on-device deployment, confirming that transition-aware triplet training yields a single reconfigurable, partitionable model for convolutional backbones as well.
  

%% file: Chapters/Conclusions.tex
\section{Limitations} 
\label{s:limitations}
EV-JEPA offers a practical solution for deploying models across heterogeneous inference infrastructures, both for standalone execution and for cross-device collaboration. However, it comes with three limitations that are worth noting.
First, EV-JEPA is tied to a teacher, as training requires a pretrained V-JEPA model
to provide the target space. Whether the stability that a frozen teacher
provides across widths (Section~\ref{ss:vjepa_distillation}) can be recovered without an
external teacher, for instance, through an EMA scheme adapted to the elastic
regime, is left to future work. Second, our elasticity is confined to width. The
depth $L$ is fixed for every standalone configuration, including the smallest,
so the narrowest subnetwork still pays the full sequential cost of $L$ layers.
Extending T3 to depth is not immediate,  since the split layer $k$ that defines a hybrid path is itself indexed by depth.
Third, T3 operates over a fixed set of widths, and the cost of training grows
with that set. Supporting a broader range of devices, therefore, increases training time. Future work will address these specific problems to make EV-JEPA and the next generation of T3 even more practical in real-world scenarios.

\section{Conclusions}\label{s:conclusion}
We introduced EV-JEPA, a single self-supervised video model that is at once
  width-elastic and natively partitionable across a heterogeneous edge-cloud
  continuum. Its core enabler is Transition-Aware Triplet Training (T3),
  which at every step optimizes a width pair both as two standalone
  configurations and as the endpoints of the hybrid path connecting them,
  making cross-width compatibility a trained property rather than an emergent
  one, and doing so under a dense predictive objective for the first time. A
  single trained model spans a wide capacity range: across SSv2, K400, and
  EK100 action anticipation, T3 stays within 1 accuracy points of independently trained
  standalone baselines and improves over prior elastic training by up to 11 points of
  at intermediate widths, while resolving the intermediate-subnetwork
  degradation of earlier approaches. Beyond standalone use, hybrid
  cross-device execution lets a small on-device encoder be completed at full
  width on a stronger device, recovering to within 1.1 points of the
  highest-capacity standalone model from an on-device $w{=}192$ model,
  and supporting many-to-one, any-to-any, and variable-split configurations.
  At full width, the same model matches or exceeds fixed-capacity
  state-of-the-art encoders (75.4 SSv2, 81.0 average) at roughly a third of
  their parameters. Finally, the elastic principle extends beyond the
  backbone: to an elastic token compressor that cuts transmitted payload by
  up to $512\times$, to vision-language transfer under VL-JEPA, and to
  convolutional architectures. Together, these results show that a single
  model, trained once, can serve the entire device continuum with
  reconfigurable compute and collaborative execution, rather than a family of
  rigid, per-device networks.